%% file: main.tex
\documentclass[10pt,twocolumn,letterpaper]{article}

\usepackage[pagenumbers]{cvpr}

\input{preamble}

\definecolor{cvprblue}{rgb}{0.21,0.49,0.74}
\usepackage[pagebackref,breaklinks,colorlinks,allcolors=cvprblue]{hyperref}

\hypersetup{pdftitle={SCOPE: Single-Cell Omics Prediction from Histology via RNA-Protein Prior Enhancement},pdfauthor={Leyang Tao, Xuan Gong, Luyuan Xie, Arjun Beniwal, Yuqi Tan, Kun-Hsing Yu}}
\def\confName{CVPR}
\def\confYear{2027}

\title{SCOPE: Single-Cell Omics Prediction from Histology via RNA–Protein Prior Enhancement}

\author{
Leyang Tao$^{1}$\quad Xuan Gong$^{1}$\quad Luyuan Xie$^{2}$\quad Arjun Beniwal$^{1}$\quad Yuqi Tan$^{3}$\quad Kun-Hsing Yu$^{1,*}$\\[0.3em]
{\normalsize $^{1}$Department of Biomedical Informatics, Harvard Medical School, Boston, MA, USA}\\
{\normalsize $^{2}$Beijing University of Civil Engineering and Architecture, Beijing, China}\\
{\normalsize $^{3}$Department of Microbiology and Immunology, Stanford University, Stanford, CA, USA}\\
{\normalsize $^{*}$Correspondence to Kun-Hsing Yu (\texttt{Kun-Hsing\_Yu@hms.harvard.edu})}
}

\begin{document}
\maketitle
\input{sec/0_abstract}
\input{sec/1_intro}

\input{sec/2_related}

\input{sec/4_method}

\input{sec/5_experiments}
\input{sec/6_conclusion}

\noindent\textbf{Use of LLMs.} We used Claude (Anthropic) to assist with coding and drafting. The authors verified all text, numbers, tables and figures.

{
    \small
    \bibliographystyle{ieeenat_fullname}
    \bibliography{main}
}

\end{document}

%% file: preamble.tex
\usepackage{booktabs}
\usepackage{multirow}
\usepackage{amsmath,amssymb}
\usepackage{bm}
\usepackage{xspace}
\usepackage{colortbl}
\usepackage{graphicx}
\newcommand{\method}{SCOPE\xspace}
\newcommand{\hl}[1]{\textbf{#1}}
\newcommand{\second}[1]{\underline{#1}}
\newcommand{\third}[1]{\textit{#1}}
\newcommand{\HE}{H\&E\xspace}

%% file: sec/0_abstract.tex
\begin{abstract}

Predicting the RNA and protein content of cells from routine \HE images could extend spatial molecular profiling to existing slide archives. However, both \HE{}--spatial proteomics and \HE{}--spatial transcriptomics paired datasets are limited by the number of available markers and the insufficient availability of high-quality, cell-resolved data. We present \method (\textbf{S}ingle-\textbf{C}ell \textbf{O}mics \textbf{P}rediction via RNA--protein prior \textbf{E}nhancement), a unified framework that leverages biologically complementary information between RNA and proteins to address the limited availability of single-omics supervision. Our key insight is that, although RNA and proteins represent distinct molecular modalities, their biological relationships provide valuable shared knowledge that can benefit single-omics prediction tasks. To this end, \method introduces biologically informed biomarker priors to enhance shared gene-level representations, enabling knowledge transfer across RNA and protein modalities through gene–protein correspondence and molecular associations. By exploiting these complementary relationships, \method enables both prediction tasks to benefit from a shared backbone enriched with cross-omics biological knowledge. Furthermore, its biomarker correlation design facilitates the prediction of unseen markers, extending spatially resolved molecular profiling beyond directly supervised targets. Across test datasets spanning 16 organs,  \method improves the average RNA Pearson correlation over the strongest baseline by 8.8\%. On two
  external protein cohorts, it improves the average Pearson correlation over the strongest
  baseline by 9.7\% for seen markers and 11.9\% for unseen markers.
\end{abstract}

%% file: sec/1_intro.tex
\section{Introduction}
\label{sec:intro}

\begin{figure}[t]
  \centering
  \includegraphics[width=\linewidth]{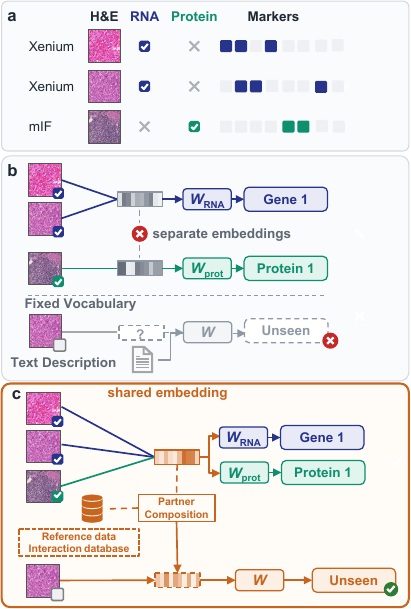}
  \caption{\textbf{\method versus existing \HE models.} 
   (a) Spatial RNA and protein assays measure different marker panels across tissue sections. (b) Existing approaches give RNA and protein separate embeddings. Fixed-vocabulary models predict a fixed list of output channels, and query-based methods read pretrained marker representations. (c) \method reads RNA and protein through one embedding per gene, so a protein without protein labels is read through the RNA-trained embedding of its gene, and each embedding adds a combination of related genes chosen by single-cell co-expression and interaction databases.}
  \label{fig:task}
\end{figure}

Spatially resolved transcriptomic and proteomic measurements provide complementary insights into cellular states and tissue organization in health and disease~\cite{rao2021exploring,hickey2022spatial,dezem2024spatially}. However, obtaining these measurements at single-cell resolution requires costly and technically demanding assays, including in situ spatial transcriptomics~\cite{janesick2023xenium} and multiplex immunofluorescence (mIF) techniques such as CODEX~\cite{goltsev2018codex}. In contrast, hematoxylin and eosin (\HE) staining is routinely performed in pathology. Predicting single-cell RNA and protein expression directly from \HE images could therefore extend spatial molecular profiling to routinely collected tissue sections and existing slide archives, supporting biological investigation and potentially informing clinical decisions.

Recent models have demonstrated the feasibility of predicting spatial molecular profiles from histology. However, many approaches target a single molecular modality or a predefined set of markers~\cite{he2020stnet,gao2026cello,valanarasu2026gigatime}, limiting their applicability across assays with different molecular panels (\cref{fig:task}(b)). Joint learning from transcriptomic and proteomic data offers a promising opportunity to leverage their complementary biological information. However, these modalities are rarely profiled on the same tissue sections, while variations in marker panels leave many molecular targets without direct \HE supervision. These limitations motivate a unified framework that can exploit complementary knowledge across unpaired omics modalities and generalize to molecular markers beyond those directly observed during training.

Addressing these challenges requires a unified representation that bridges complementary supervision across molecular modalities and biomarker targets. The correspondence between proteins and their coding genes provides a natural biological link between RNA and protein measurements, enabling supervision from either modality to contribute to shared gene-level representations. However, such correspondence alone cannot address genes absent from the training panels. To further extend supervision beyond directly observed targets, biological relationships among genes provide complementary priors that facilitate knowledge transfer from supervised to unseen genes. Motivated by these insights, we develop a biologically informed shared gene representation that integrates cross-omics supervision with biomarker-level biological relationships, enabling knowledge transfer across both modalities and molecular targets.

In this paper, we introduce \method, a unified framework for single-cell RNA and protein prediction from \HE images.  At its core, \method leverages the biological correspondence between genes and their encoded proteins to establish a shared molecular representation, enabling complementary supervision across transcriptomic and proteomic modalities.  To extend prediction beyond directly supervised targets, we further introduce biologically informed biomarker co-encoding, which incorporates gene co-expression patterns, molecular interactions, and pathway relationships to transfer knowledge from observed to unseen markers.
In addition, \method adopts a dual-branch architecture that integrates cell-level molecular prediction with dense spatial expression modeling, leveraging complementary supervision from subcellular transcript localization and protein image intensities. A fusion strategy further combines predictions from the two branches to enable comprehensive spatially resolved molecular profiling. Our contributions are summarized below.
\begin{itemize}
    \item We introduce SCOPE, a unified framework for
    single-cell RNA and protein prediction from H\&E
    images, enabling joint learning from spatial transcriptomic and proteomic data acquired on separate tissue sections through shared gene-level representations.

    \item We propose biologically informed biomarker co-encoding, leveraging gene relationships and single-cell co-expression patterns to enable prediction of molecular markers without direct H\&E supervision, extending spatial molecular profiling beyond the markers observed during training.

  \item We evaluate \method across 16 organs and two external protein cohorts. \method improves the
  average Pearson correlation over the strongest RNA baseline by 8.8\%, and over the strongest
  protein baseline by 9.7\% and 11.9\% for seen and unseen markers, respectively.
\end{itemize}

%% file: sec/2_related.tex
\section{Related work}
\label{sec:related}
\noindent\textbf{Molecular profiles from \HE.}
Most methods predict either spatial RNA or protein expression from \HE, at the cell, spot, or pixel level. For RNA, DeepSpot-M uses gene-conditioned decoding from image tokens~\cite{nonchev2026deepspotm}, and CELLO predicts per-cell expression using features from a pathology foundation model~\cite{gao2026cello}. Pixel-level approaches reconstruct expression maps from spot-level or binned-transcript supervision~\cite{zhang2024istar,yao2026pixel2gene}. For protein, virtual staining methods such as ROSIE, GigaTIME, and HistoPlexer generate multiplexed images~\cite{wu2025rosie,valanarasu2026gigatime,andani2025histoplexer}. Fewer methods address both assays. H2O uses parallel decoders trained on consecutive sections profiled by both assays~\cite{gu2026h2o}; CUBE fuses protein-derived and predicted transcriptomic representations~\cite{ge2026cube}. \method trains on sections with either RNA or protein measurements from unpaired datasets, but couples the assays through shared gene embeddings and predicts all requested targets in one forward pass. It combines cell-level and dense expression predictions.

\noindent\textbf{Marker embeddings.}
Models predicting many molecular targets use learned identifiers or embeddings derived from sequence or text. CAPTAIN and CARE maintain separate gene and protein embeddings~\cite{ji2026captain,zhang2026care}, whereas UCE and VirTues use protein language-model representations~\cite{rosen2026uce,wenckstern2026virtues}. To generalize beyond directly supervised genes, SGN uses gene descriptions~\cite{yang2024sgn}, DeepSpot-M uses frozen gene embeddings~\cite{nonchev2026deepspotm}, GSAT-RAG retrieves expression information from seen genes~\cite{wu2026gsatrag}, and HINGE conditions a pretrained single-cell model on histology~\cite{fang2026hinge}. In contrast, \method learns a shared embedding for each gene and its encoded protein from both assays. For genes without \HE-linked labels, it complements this embedding with a sparse, signed combination of related genes, chosen by single-cell co-expression and by interaction and pathway resources and weighted by single-cell expression covariation.

%% file: sec/4_method.tex
\section{Method}
\label{sec:method}

\begin{figure*}[t]
  \centering
  \includegraphics[width=\textwidth]{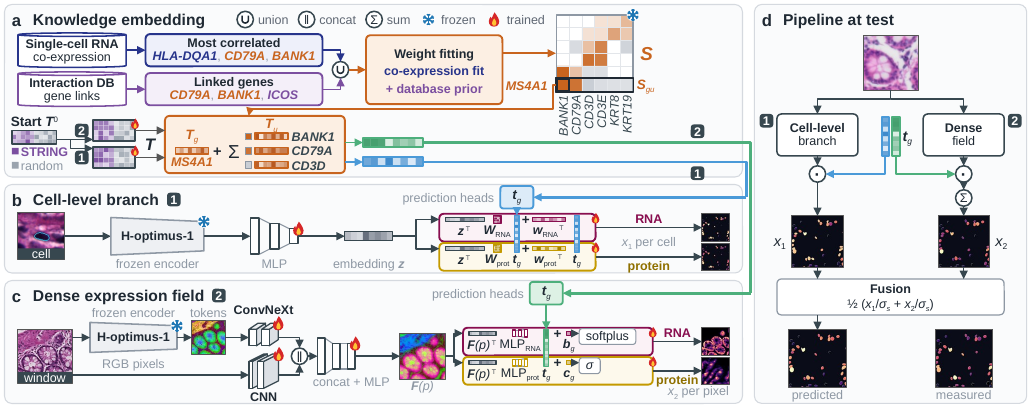}
  \caption{\textbf{Overview of \method.}
 (a)~Knowledge embedding. Single-cell RNA and interaction databases select the partners of each gene and fit their weights
  $\mathbf{S}$. Each branch (tags b and c) learns its own embedding table $\mathbf{T}$ from the shared start $\mathbf{T}^0$
  and reads gene $g$ through $\mathbf{t}_g=\mathbf{T}_g+\mathbf{S}_g\mathbf{T}$, which both its RNA and protein heads use.
  (b)~The cell-level branch scores the embedding of each nucleus against $\mathbf{t}_g$.
  (c)~The dense expression field scores each pixel, computed from H-optimus-1 tokens and the RGB image, against $\mathbf{t}_g$.
  (d)~At test time, only the \HE image and its nuclei are used, and the two branches are averaged with equal weights.}
  \label{fig:pipeline}
\end{figure*}

\noindent\textbf{Method overview.} \method predicts the RNA and protein of every \HE nucleus from the image alone
  (\cref{fig:pipeline}). Every marker is read through a gene embedding that RNA and protein share and that adds a
  combination of related genes (\cref{sec:rows}). Two branches read these embeddings at two resolutions, a cell model that
  scores one image vector per nucleus (\cref{sec:cell}) and a dense expression field that scores every pixel and
  is trained on transcript positions and protein images (\cref{sec:map}). Late fusion averages their predictions
  (\cref{sec:mix}).

\subsection{Biomarker co-encoding}
\label{sec:rows}
We compose each embedding from the gene's own learnable vector and the vectors of related genes, allowing information learned from supervised partner genes to improve the representations of genes with limited or no direct supervision. We take candidate partners of each gene from single-cell co-expression and from STRING~\cite{szklarczyk2023string}, which
  scores functional associations between proteins, and Reactome~\cite{milacic2024reactome}, which organizes curated
  molecular reactions into pathways. The signed weight of each partner is estimated from gene covariation in single-cell
  transcriptomic data, and the database links set the prior mean of this estimate. Both branches query marker $g$ through its embedding in
\begin{equation}
  \mathbf{t}=\mathbf{T}+\mathbf{S}\mathbf{T},\qquad
  \mathbf{t}_g=\mathbf{T}_g+\textstyle\sum_{u}S_{gu}\mathbf{T}_u ,
  \label{eq:rows}
\end{equation}
with learnable embeddings $\mathbf{T}\in\mathbb{R}^{V\times1024}$ for the $V=19{,}496$ genes and protein-marker
entries, and relation weights $\mathbf{S}\in\mathbb{R}^{V\times V}$, the signed adjacency of a gene graph
(\cref{eq:rows} is one propagation step).

\noindent\textbf{Learnable embeddings.}
Embedding $\mathbf{T}_g$ starts at
$\mathbf{T}^{0}_g=[\,\text{STRING }512\,|\,\text{random }512\,]$, a spectral embedding (normalized
Laplacian~\cite{belkin2002laplacian}, as in~\cite{tang2023scmoformer}) of the STRING v12 network and a fixed Gaussian vector. An entry with no STRING protein takes a random first
half of the same norm. A loss that reads $g$ trains $\mathbf{T}_g$ and,
through $\mathbf{S}$, the embeddings of $g$'s partners. An embedding that no loss reads gets no gradient and no
weight decay, so it stays at $\mathbf{T}^{0}_g$. Such a gene thus reads $\mathbf{T}^{0}_g+\mathbf{S}_g\mathbf{T}$
in both the cell model and the dense branch, its start embedding plus its partners' trained embeddings. Each part trains its own
$\mathbf{T}$, and both share $\mathbf{S}$.

\noindent\textbf{Relation weights.}
 $\mathbf{S}_g$ holds a signed weight for each candidate partner of $g$. The candidates $\mathcal{A}_g$ are the genes most
  correlated with $g$ in single-cell RNA and the genes among the strongest links of $g$ in STRING or Reactome (through a
  reaction, as input, output, catalyst, or regulator, or a complex), drawn from genes measured on many training
  sections. Ridge regression toward a prior mean $\mathbf{m}_g$ built from the database links gives the weights
\begin{equation}
  \begin{split}
  \mathbf{S}_{g,\mathcal{A}_g}={}&\Big(\sum_{d}\mathbf{Z}_{d}^{\top}\mathbf{Z}_{d}+(\alpha+\omega)\mathbf{I}\Big)^{-1}\\
  &\Big(\sum_{d}\mathbf{Z}_{d}^{\top}\mathbf{r}_{d}+\omega\,\mathbf{m}_g\Big),\quad \alpha=1,\ \omega=10^{4},
  \end{split}
  \label{eq:ridge}
\end{equation}
  
For each dataset $d$, $\mathbf{Z}_d$ contains the
$\ln(1+c)$ expression of candidate partners in
$\mathcal{A}_g$, and $\mathbf{r}_d$ contains that of
target gene $g$, both z-scored within the dataset.
Candidates not measured in $d$ are assigned zero.
We use the training Xenium sections and RNA profiles
from the training CITE-seq datasets~\cite{stoeckius2017citeseq}.

We estimate partner weights using ridge regression
with a biological prior $\mathbf{m}_g$.
The penalty $\alpha$ stabilizes the estimates,
while $\omega$ controls their shrinkage toward the
prior. Since expression values are standardized,
$\omega$ approximately determines the prior's
strength relative to the number of observed cells.
For each linked partner, the prior combines a constant,
its STRING association score, and indicators of
a shared Reactome reaction or protein complex.
Unlinked partners receive a zero prior.
The coefficients are fitted by least squares using
genes from other folds of a random gene partition,
so a gene's own data are excluded when fitting
its prior coefficients.
This allows biological relationships to guide
the weights when expression evidence is limited,
while observed co-expression has greater influence
when sufficient data are available.

We estimate these weights for genes in our Xenium
panels.
For other genes, $\mathbf{S}_g=\mathbf{0}$,
giving $\mathbf{t}_g=\mathbf{T}_g$.
Each protein marker inherits the weights of its
coding gene $\gamma(g)$.

\subsection{Cell branch}
\label{sec:cell}
The cell-level branch predicts the expression of each nucleus from one frozen image vector. One MLP and one set of gene embeddings serve both assays, and each assay has its own prediction head.
Nucleus $i$ is described by one frozen H-optimus-1~\cite{scalbert2026hoptimus1} vector
$\mathbf{u}_i\in\mathbb{R}^{1536}$, the average of the patch tokens of a 224-pixel tile at
0.5\,\textmu m/px, weighted by the fraction of each token that the nucleus covers. An MLP
(LayerNorm, $1536\!\to\!1024$, GELU, $1024\!\to\!512$) gives $\mathbf{z}_i\in\mathbb{R}^{512}$, and a
bilinear prediction head gives marker $g$ of assay $a$:
\begin{equation}
  x_1(i,g)=\mathbf{z}_i^{\top}\mathbf{W}_a\,\mathbf{t}_g+\mathbf{w}_a^{\top}\mathbf{t}_g ,
  \label{eq:readout}
\end{equation}
where $\mathbf{t}_g\in\mathbb{R}^{1024}$ is the embedding of $g$ (\cref{sec:rows}). A protein marker $g$ reads the
  embedding $\mathbf{t}_{\gamma(g)}$ of $\gamma(g)$, the gene that encodes its target. Two weights are trained per assay: $\mathbf{W}_a\in\mathbb{R}^{512\times1024}$ couples the
cell's state with the marker, and $\mathbf{w}_a\in\mathbb{R}^{1024}$ gives the marker's baseline level
$\mathbf{w}_a^{\top}\mathbf{t}_g$, shared by all cells. The target $y(i,g)$ is the measured value, $\ln(1+c)$ of the transcript count $c$ for RNA and $\ln(1+\bar{I})$ of the mean
  intensity $\bar{I}$ for protein, z-scored per marker with training statistics. Over the step's cells
  $\mathcal{B}$, all from one training section $s$, and the markers $\mathcal{P}_s$ that $s$ measures, the loss is the mean
  squared error:
\begin{equation}
  \mathcal{L}_{\mathrm{cell}}=\frac{1}{|\mathcal{B}|\,|\mathcal{P}_s|}\sum_{i\in\mathcal{B}}\sum_{g\in\mathcal{P}_s}
  \big(x_1(i,g)-y(i,g)\big)^2 .
  \label{eq:mse}
\end{equation}

\subsection{Dense branch}
\label{sec:map}
 The dense branch predicts expression on a pixel grid, trained on transcript positions and CODEX pixels, and a cell takes the sum
  (RNA) or mean (protein) of the field over its pixels.
  It reads the \HE at two scales. Foundation-model tokens carry tissue context but are coarse, one 7\,\textmu m token being
  about the width of a nucleus, while the pixels resolve nuclear outlines.
  The field is predicted on 0.875\,\textmu m pixels, in tiles of 56\,\textmu m ($64\times64$ pixels).
  Frozen H-optimus-1 reads the 112\,\textmu m box around each tile at 0.5\,\textmu m/px
  ($16\times16$ tokens of 7\,\textmu m). We keep the $8\times8$ tokens over the tile, one per
  $8\times8$ pixels, so that every kept token is encoded with at least 28\,\textmu m of surrounding tissue, and reduce them
  to 256 dimensions by a PCA fitted on training tiles. On windows of $4\times4$ tiles
  (224\,\textmu m, $32\times32$ tokens, $256\times256$ pixels), a $1\times1$ convolution and two
  ConvNeXt blocks~\cite{liu2022convnext} process the tokens, whose $7\times7$ depthwise convolutions let each token use its
  neighbors within 42\,\textmu m, across tiles that the encoder saw separately. The output is upsampled $\times8$ bilinearly
  and concatenated with a two-layer CNN ($3\times3$ kernels) on the \HE at 0.875\,\textmu m/px, which adds the local detail
  a token cannot place, such as where a nucleus ends. A per-pixel MLP gives the field $\mathbf{F}(p)\in\mathbb{R}^{256}$ at
  pixel $p$. These layers hold 1.3M weights. The RNA and protein heads below both read this one field, so both losses train
  them.

\noindent\textbf{RNA head.}
The density of gene $g$ at pixel $p$ of section $s$, in transcripts per \textmu m$^2$, is
$\lambda(p,g)=\eta_s\,\mathrm{softplus}(\mathbf{F}(p)^{\top}\mathbf{q}_g+b_g)$. The query
$\mathbf{q}_g=\mathrm{MLP}_{\mathrm{RNA}}(\mathbf{t}_g)\in\mathbb{R}^{256}$ maps the gene's embedding into the space of the
field, so $\mathbf{F}(p)^{\top}\mathbf{q}_g$ scores how strongly pixel $p$ shows gene $g$. $b_g$ is the gene's baseline abundance and $\eta_s>0$ the detection efficiency of training section $s$, a technical factor set by chemistry,
gene panel and tissue preparation. The loss is the Poisson negative log-likelihood, \ie the point-process likelihood of
the transcript positions \cite{walter2023fishfactor} at pixel resolution:
\begin{equation}
  \mathcal{L}_{\mathrm{RNA}}=\frac{1}{N}\sum_{p}\sum_{g\in\mathcal{P}_s}
  \big[\mu(p,g)-n(p,g)\log\mu(p,g)\big],
  \label{eq:poisson}
\end{equation}
where $\mu(p,g)=\lambda(p,g)\,\Delta A$ is the expected count, $\Delta A=0.875^2$\,\textmu m$^2$ is the area of one pixel, $n(p,g)$ is the number of transcripts of $g$ at pixel $p$, $p$
runs over the pixels of the stored tiles of the step's Xenium windows, and $N$ counts their transcripts. Unlike the cell model, the dense branch counts every transcript, whether or not it lies in a segmented cell.

\noindent\textbf{Protein head.}
The intensity of protein marker $g$ is $\pi(p,g)=\sigma(\mathbf{F}(p)^{\top}\mathbf{q}'_g+c_g)$, with the logistic
function $\sigma$ keeping it between 0 and 1, like the target. The query $\mathbf{q}'_g=\mathrm{MLP}_{\mathrm{prot}}(\mathbf{t}_{\gamma(g)})\in\mathbb{R}^{256}$ maps the embedding of the marker's gene into the space of the field with weights of its own, and $c_g$ is the marker's baseline intensity. Its target is
$v(p,g)=\min(\bar{I}(p,g)/h_g,1)$: the mean raw intensity of the CODEX pixels registered into $p$,
divided by the marker's saturation level $h_g$, its 99.9th percentile over training pixels~\cite{balezo2025miphei}. The loss is the binary cross-entropy
\begin{equation}
  \mathcal{L}_{\mathrm{prot}}=-\frac{1}{M}\sum_{p}\sum_{g\in\mathcal{P}_s}
  \big[v\log\pi+(1-v)\log(1-\pi)\big],
  \label{eq:bce}
\end{equation}
with $v$ and $\pi$ taken at $(p,g)$, $p$ running over the pixels with CODEX data in the step's CODEX
windows, and $M$ is the number of pairs of pixels and markers.

\noindent\textbf{Pooling per cell.}
  The dense branch reads over each cell's footprint $\Omega_i$: the pixels whose center lies inside the \HE nucleus $i$, the same mask over which the cell vector $\mathbf{u}_i$ is pooled, so that the read-out
  does not reach into neighboring cells. For protein, the nucleus is first grown by 1.5\,\textmu m, the 3-pixel growth with which the
  CODEX targets are quantified, and a pixel that two grown nuclei reach goes
  to the nearer one. The dense branch gives cell $i$ a sum for RNA and a mean for protein:
\begin{equation}
  \begin{aligned}
    x_2(i,g)&=\ln\!\Big(1+\Delta A\sum_{p\in\Omega_i}\bar{\lambda}(p,g)\Big) &&\text{(RNA)},\\
    x_2(i,g)&=\frac{1}{|\Omega_i|}\sum_{p\in\Omega_i}\pi(p,g) &&\text{(protein)},
  \end{aligned}
  \label{eq:readmap}
\end{equation}
where $\bar{\lambda}$ is $\lambda$ with $\eta_s$ replaced by $\bar{\eta}$, the median detection efficiency of
the training sections.

\noindent\textbf{Per-gene late fusion.}
\label{sec:mix}
The cell branch and the dense branch see the \HE at two scales, so their errors differ in part.
We combine them without fitting anything. For cell $i$ and marker $g$, the cell model predicts $x_1(i,g)$ (\cref{eq:readout}) and the dense branch gives
$x_2(i,g)$ (\cref{eq:readmap}). In test section $s$, the final prediction averages the two after each is divided by its standard deviation,
\begin{equation}
  \hat{x}(i,g)=\frac{1}{2}\left(\frac{x_1(i,g)}{\sigma_s(x_1(\cdot,g))}
  +\frac{x_2(i,g)}{\sigma_s(x_2(\cdot,g))}\right),
  \label{eq:mix}
\end{equation}
where $\sigma_s(x_k(\cdot,g))$ is the standard deviation of branch $k$'s predictions for marker $g$ over the cells of $s$. It needs
no labels, so it is available on a test set. The weights are equal for every marker and both assays.

%% file: sec/5_experiments.tex
\section{Experiments}
\label{sec:experiments}

\subsection{Setup}
\label{sec:exp_setup}
\input{tables/tab_rna}
\input{tables/tab_protein}
\input{tables/tab_ablation}

\noindent\textbf{Data.}
We train on 43 Xenium sections from 10 organs (11 in-house and 32 public), with panels of 280--5,098 genes (median 300), and 76 CODEX in-house
sections (517 intestinal tissue pieces) from three donors,
with 58 or 59 protein markers. Sections are registered
to \HE using VALIS~\cite{gatenbee2023valis}, and
measured cells are matched one-to-one with \HE nuclei,
yielding 7.87M Xenium and 26.49M CODEX training cells.
The Xenium sections cover 5,828 genes, each measured
on a median of three sections. To estimate the relation
weights $\mathbf{S}$, we additionally use RNA profiles
of 2.44M cells from 150 training CITE-seq datasets
across 21 studies~\cite{stoeckius2017citeseq}.
These datasets provide gene co-expression information
but no H\&E-linked supervision. Four Xenium and three
CODEX sections are reserved for validation.

We evaluate on (i) 16 Xenium sections from 16 organs,
six of them from organs without any training section (3.38M cells, 5,649 genes);
(ii) 22 CODEX sections from a fourth donor, in-house colon
(4.11M cells, 57 markers); and (iii) two external
colorectal cancer cohorts: CRC~\cite{schurch2020crc,
schurch2020crcdata} (35 patients, 139 cores, 192,779 cells)
and ORION~\cite{lin2023orion} (40 patients, 41 slides,
22.32M cells). No evaluation patient appears in
SCOPE's training data. Seen RNA genes are training
targets, while unseen genes are withheld from direct
training supervision. Protein markers are considered
unseen if they are absent from the CODEX training panels.

\noindent\textbf{Evaluation.}
We compute Pearson correlation between predicted and
measured expression across cells for each marker,
then average over markers and evaluation units.
Missing predictions contribute zero correlation.
For the baseline comparison in \cref{tab:protein}, CRC
correlations are computed using cells pooled by patient
across 135 eligible cores (190,666 cells), and ORION
is evaluated on the four patients excluded from
MIPHEI-ViT training. Component ablations use all
ORION patients.

Statistical significance is assessed using one-sided
paired $t$-tests, with Benjamini--Hochberg correction
within each table ($q<0.05$). The testing units are
section--gene pairs for \cref{tab:rna}, markers for
\cref{tab:protein}, and sections or patients for
the component ablations. Results are averaged over
three seeds.

\subsection{Implementation details}
\noindent\textbf{Training on both assays.}
\label{sec:joint}
  Both branches are trained with AdamW~\cite{loshchilov2019adamw}, and we evaluate their final checkpoints.
  The cell model interleaves the assays across steps. Each step takes the cells of one RNA or one protein section, and both
  assays update the shared MLP and gene embeddings, each with its own AdamW state~\cite{yang2023adatask}. It trains for
  120{,}000 steps, about 9 epochs over the Xenium cells and 2 over the CODEX cells (2\,h on one A40 GPU).
  The dense branch instead sees both assays in every step, four Xenium and four CODEX windows, and trains for 30{,}000
  steps, about 2 epochs over the Xenium tiles and 0.5 over the CODEX tiles (2.6--3.7\,h on one A40 or A100 GPU). Sections and CODEX tissue pieces are drawn in proportion to their number of tiles. To balance the differently scaled Poisson and cross-entropy losses (\cref{eq:poisson,eq:bce}), we divide each by its exponential moving average (decay 0.99), giving $\mathcal{L}_{\mathrm{RNA}}/\bar{\mathcal{L}}_{\mathrm{RNA}}+\mathcal{L}_{\mathrm{prot}}/\bar{\mathcal{L}}_{\mathrm{prot}}$. Unlike the cell model, the dense branch keeps one optimizer state for both assays, and its single-modality counterparts keep only the corresponding head and loss.
\subsection{Results}
\label{sec:exp_baselines}

\noindent\textbf{RNA.}
\Cref{tab:rna} compares \method with four published models on the 16 test sections. \method achieves the highest correlation in all ten columns. Averaged over the four organ groups, it reaches 0.484 on HVG10 and 0.167 on all genes, against 0.437 and 0.154 for the strongest baseline, DeepSpot-M fine-tuned on our training sections, and it is significantly above every baseline in both average columns. The margin over the models run with their released weights is much larger (CELLO 0.307 and 0.114, DeepSpot-M 0.189 and 0.049), and the two models that learn from spots, sCellST and DeepSpot2Cell, average below 0.27 on HVG10 after retraining on our sections, since a spot label sums the expression of many cells and does not tell which cell expresses a gene. The largest gains are on HVG10, the genes that vary most between the cells of a section and therefore separate its cell types and states.
The lead holds where training data are scarce. In post hoc views, \method is significantly above the fine-tuned DeepSpot-M on the six sections from organs without any training section (+0.0132), as on the ten sections from organs with one (+0.0138), and on genes labeled on only one to three training sections (+0.0193). This shows that \method transfers to organs without training sections and to genes labeled on few sections.

\noindent\textbf{Protein.}
\Cref{tab:protein} compares SCOPE with four published models across three protein test sets. Since baselines predict fixed marker panels, markers without predictions are counted as zero in the full-panel averages. SCOPE ranks first in every CRC and in-house colon column, achieving correlations of 0.278 and 0.127 on seen and unseen CRC markers, and 0.584 on in-house colon, compared with at most 0.087, 0.060, and 0.075 for the baselines. On ORION, MIPHEI-ViT achieves the highest performance but was trained on 37 of the 41 evaluation slides. The remaining four slides used in the test are in-distribution data for it. SCOPE ranks first among models not trained on ORION (0.449 seen, 0.356 unseen). Unlike fixed-output baselines, SCOPE predicts every marker in the evaluation panels, including proteins absent from its CODEX training data. These results demonstrate transfer to independent patient cohorts and prediction beyond the training protein panel.

\subsection{Ablation study on joint training}
\label{sec:exp_joint_data}
\Cref{tab:ablation} tests whether RNA supervision improves
protein prediction when the two assays are measured on
separate sections. Each part of \method is trained on
both assays and compared with a single-modality
counterpart using the same number of protein steps,
controlling for additional protein training.
Joint training improves prediction of seen protein
markers in both external cohorts. The cell model gains
+0.0271 on CRC and +0.0246 on ORION, the dense branch +0.0209
and +0.0373, and the dual branch +0.0244 and +0.0288,
all significant. The benefit extends to unseen markers,
where the dual branch gains +0.0413 on CRC and +0.0306 on
ORION, both significant, while the dense branch achieves the
largest gain on unseen CRC markers (+0.0698).

Two mechanisms may contribute to these gains.
First, RNA labels come from 43 sections across 10
organs, whereas protein labels come from intestinal
tissue of three donors, exposing the shared image
representations to a broader range of tissues.
Second, shared gene embeddings provide a route
for cross-assay transfer. All unseen protein markers
but one have coding genes supervised by RNA labels,
so their embeddings can benefit from RNA supervision.
These results demonstrate that RNA supervision
benefits protein prediction, including markers
without direct protein supervision.

 \subsection{Ablation study on dual-branch fusion}
\label{sec:exp_joint_model}
The cell and dense branches predict molecular expression
at different spatial resolutions, and the dual branch
combines their predictions with fixed equal weights.
In \cref{tab:ablation}, the dual branch significantly
outperforms the cell branch in every column and the
dense branch in every seen-marker column, by +0.0120
on RNA, +0.0099 on CRC, and +0.0144 on ORION.
These improvements suggest that the two branches
provide complementary predictive information.

For unseen markers, the relative contributions of
the branches differ. On unseen CRC markers, the
dense branch achieves the highest correlation (0.225),
followed by the dual branch (0.185) and the cell
branch (0.066). On unseen ORION markers, the dual
and dense branches achieve comparable performance
(0.386), while on unseen RNA genes, the dual branch
outperforms both individual branches.

Overall, these results show that the cell and dense
branches provide complementary predictions, with
their equal-weight combination improving performance
over both branches on seen markers. For unseen
markers, the relative contributions vary across
evaluation settings, with the dense branch providing
stronger predictions in some cases.

\subsection{Ablation study on gene embeddings}
\label{sec:exp_knowledge}

\input{tables/tab_knowledge}
\noindent\textbf{Knowledge components.}
The gene embedding
$\mathbf{t}_g=\mathbf{T}_g+\mathbf{S}_g\mathbf{T}$
(\cref{eq:rows}) incorporates two sources of knowledge:
the initialization $\mathbf{T}^0$, whose first half
encodes the gene's position in the STRING network,
and the relation weights $\mathbf{S}$, which combine
embeddings of co-expressed and biologically linked
partners. \Cref{tab:knowledge} ablates either or both
components and retrains \method. We evaluate on
98 withheld RNA genes that have relation weights and on
unseen protein markers, none of which receives direct supervision.

Both components improve withheld RNA prediction.
The dual branch rises from 0.040 without knowledge
to 0.083 with a random start and $\mathbf{S}$, 0.105 with
$\mathbf{T}^0$ alone, and 0.129 with both.
\method significantly outperforms all three ablations.
On unseen CRC proteins, \method also achieves the
highest correlation (0.185), compared with 0.152
without knowledge, while on ORION it reaches
0.386 versus 0.176.

The two components contribute differently.
$\mathbf{T}^0$ accounts for most of the improvement
on unseen proteins, which are predicted through
their coding genes' embeddings. Without $\mathbf{S}$,
the dual branch reaches 0.175 on CRC and 0.387
on ORION, compared with 0.185 and 0.386 for
the full model.

$\mathbf{S}$ helps most where a gene has no RNA label, adding +0.024 on withheld genes but +0.010 on unseen CRC proteins, whose genes (all but one) carry RNA labels. Reassigning relation weights between genes significantly lowers withheld genes from 0.129 to 0.120, so the gain depends on the right partners. $\mathbf{S}$ improves 74 of the 98 withheld genes, more for genes measured in more single-cell datasets (Spearman 0.39).

\noindent\textbf{Comparison with protein language-model embeddings.}
\label{sec:exp_esm}
\begin{figure}[t]
  \centering
  \includegraphics[width=\linewidth]{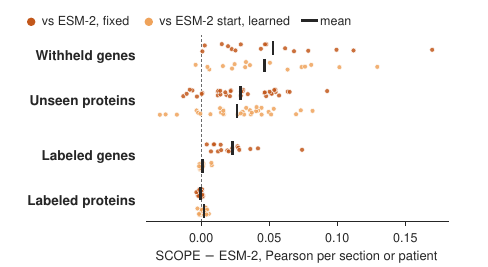}
  \caption{\textbf{\method against ESM-2 gene embeddings.} Pearson of \method minus ESM-2 per test section, or per CRC patient for unseen proteins (dots), and the mean (bar). Fixed: ESM-2 vectors as frozen gene embeddings. Start, learned: ESM-2 vectors as the start of learned embeddings.}
  \label{fig:esm2}
\end{figure}

Protein language models offer an alternative source
of gene embeddings, as used in UCE~\cite{rosen2026uce}
and VirTues~\cite{wenckstern2026virtues}.
We extract mean-pooled last-layer representations
from 15B-parameter ESM-2~\cite{lin2023esm2} for
the longest protein encoded by each gene, reduce
them to 1,024 dimensions using PCA, and match
their mean norm to our embeddings.
We evaluate these representations either frozen
or as learnable initializations, without relation
weights. Data, training steps, seeds, and
dual-branch fusion remain unchanged.

\Cref{fig:esm2} compares \method with both variants.
On withheld RNA genes, \method improves correlation
by 0.053 and 0.047 over frozen and learnable ESM-2,
respectively, and by 0.029 and 0.026 on unseen CRC
proteins. On labeled RNA genes, \method outperforms
frozen ESM-2 (+0.023) but performs comparably to
learnable ESM-2; differences on labeled in-house
colon proteins are within 0.002.

These results suggest that learnable sequence-based
embeddings can approach \method on directly
supervised targets, but do not replace the benefits
of network-based initialization and co-expression
composition for markers without direct \HE supervision.

%% file: tables/tab_rna.tex
\begin{table*}[t]
\centering
\caption{\textbf{Single-cell RNA prediction on 10x Genomics by organ group.}
Pearson correlation per gene across cells, averaged over genes
and sections within each group (section counts in parentheses).
HVG10 denotes the ten genes with the highest variance of
$\ln(1+c)$ per section; All includes all measured genes.
Genes without predictions count as zero.
Digestive: colon, stomach, liver, pancreas;
lymphoid: bone marrow, lymph node, tonsil;
reproductive and urinary: breast, ovary, cervix,
prostate, kidney cancer;
thorax, brain and skin: lung, heart, brain, skin.
Average is the mean of the four group values.
$^\dagger$ Released weights;
$^\ddagger$ fine-tuned or retrained on our training
and validation sections.
\method: mean over 3 seeds (s.d.\ $\leq 0.008$).
Bold, underline, and italic indicate the top three
values per column.}
\label{tab:rna}
\footnotesize
\setlength{\tabcolsep}{9pt}
\begin{tabular}{@{}lcccccccc@{\hspace{6pt}}cc@{}}
\toprule
 & \multicolumn{2}{c}{Digestive (4)} & \multicolumn{2}{c}{Lymphoid (3)} & \multicolumn{2}{c}{Repro., urinary (5)} & \multicolumn{2}{c}{Thorax, brain, skin (4)} & \multicolumn{2}{c}{Average} \\
\cmidrule(lr){2-3}\cmidrule(lr){4-5}\cmidrule(lr){6-7}\cmidrule(lr){8-9}\cmidrule(lr){10-11}
Model & HVG10 & All & HVG10 & All & HVG10 & All & HVG10 & All & HVG10 & All \\
\midrule
CELLO$^\dagger$~\cite{gao2026cello} & 0.454 & 0.202 & 0.216 & 0.055 & 0.281 & 0.096 & 0.278 & 0.103 & 0.307 & 0.114 \\
DeepSpot-M$^\dagger$~\cite{nonchev2026deepspotm} & 0.324 & 0.082 & 0.096 & 0.021 & 0.207 & 0.045 & 0.130 & 0.046 & 0.189 & 0.049 \\
DeepSpot-M$^\ddagger$~\cite{nonchev2026deepspotm} & \third{0.535} & \second{0.212} & 0.285 & \third{0.109} & 0.467 & \third{0.139} & \second{0.460} & \third{0.156} & 0.437 & \third{0.154} \\
sCellST$^\ddagger$~\cite{chadoutaud2026scellst} & 0.201 & 0.064 & 0.051 & 0.020 & 0.185 & 0.042 & 0.101 & 0.028 & 0.134 & 0.038 \\
DeepSpot2Cell$^\ddagger$~\cite{nonchev2025deepspot2cell} & 0.361 & 0.133 & 0.159 & 0.055 & 0.315 & 0.084 & 0.224 & 0.065 & 0.265 & 0.084 \\
\midrule
\method\ cell branch & \second{0.537} & \third{0.206} & \third{0.323} & 0.106 & \third{0.507} & 0.136 & \third{0.439} & 0.146 & \second{0.452} & 0.149 \\
\method\ dense branch & 0.510 & 0.203 & \second{0.336} & \second{0.115} & \second{0.517} & \second{0.146} & 0.426 & \second{0.158} & \third{0.447} & \second{0.155} \\
\method & \hl{0.561} & \hl{0.223} & \hl{0.363} & \hl{0.124} & \hl{0.548} & \hl{0.155} & \hl{0.464} & \hl{0.168} & \hl{0.484} & \hl{0.167} \\
\bottomrule
\end{tabular}
\end{table*}

%% file: tables/tab_protein.tex
\begin{table*}[!t]
\centering
\caption{\textbf{Protein compared with published models.} Pearson correlation per marker across cells, averaged
over markers and evaluation units.
CRC uses 135 cores (190,666 cells), pooled by patient;
ORION uses four patients excluded from MIPHEI-ViT training.
Seen and unseen indicate whether a marker receives
direct CODEX supervision during SCOPE training.
In-house colon cohort reports all 57 markers.
Markers without predictions count as zero.
{\tiny\,(x)} denotes the number of predicted markers;
-- indicates an unavailable prediction.
Avg.\ Seen averages CRC seen, ORION seen, and in-house colon all;
Avg.\ Unseen averages CRC and ORION unseen.
Col.\ IV: collagen IV; Vim.: vimentin;
E-cad.: E-cadherin.
$^\dagger$ Trained on other slides from the ORION cohort.
Gray text indicates in-distribution results.
Bold, underline, and italic indicate the top three
values per column.} 

\label{tab:protein}
\footnotesize
\setlength{\tabcolsep}{1.5pt}
\begin{tabular}{@{}lcccccc cccccc ccc cc@{}}
\toprule
 & \multicolumn{6}{c}{CRC} & \multicolumn{6}{c}{ORION} & \multicolumn{3}{c}{In-house colon cohort} & \multicolumn{2}{c}{Avg.} \\
\cmidrule(lr){2-7}\cmidrule(lr){8-13}\cmidrule(lr){14-16}\cmidrule(l){17-18}
 & \multicolumn{3}{c}{Seen (29)} & \multicolumn{3}{c}{Unseen (25)} & \multicolumn{3}{c}{Seen (11)} & \multicolumn{3}{c}{Unseen (4)} & \multicolumn{3}{c}{All (57)} & Seen & Unseen \\
\cmidrule(lr){2-4}\cmidrule(lr){5-7}\cmidrule(lr){8-10}\cmidrule(lr){11-13}\cmidrule(lr){14-16}
Model & CD45 & CD31 & Mean & GATA3 & T-bet & Mean & Pan-CK & CD45 & Mean & E-cad. & PD-1 & Mean & Col.\,IV & Vim. & Mean & & \\
\midrule
GigaTIME~\cite{valanarasu2026gigatime} & -- & -- & 0.025{\tiny\,(8)} & -- & \second{0.010} & 0.013{\tiny\,(4)} & -- & -- & 0.041{\tiny\,(3)} & -- & \third{0.139} & 0.096{\tiny\,(3)} & -- & -- & 0.010{\tiny\,(9)} & 0.025 & 0.055 \\
ROSIE~\cite{wu2025rosie} & \third{0.196} & \third{0.087} & \third{0.064}{\tiny\,(16)} & \second{$-$0.034} & -- & \second{0.060}{\tiny\,(12)} & \third{$-$0.023} & \third{0.228} & \third{0.070}{\tiny\,(9)} & \third{$-$0.060} & \second{0.153} & \third{0.125}{\tiny\,(4)} & \second{0.067} & \second{0.173} & \third{0.029}{\tiny\,(17)} & \third{0.054} & \third{0.093} \\

MIPHEI-ViT$^{\dagger}$~\cite{balezo2025miphei}
& \second{0.632}
& \second{0.436}
& \second{0.087}{\tiny\,(7)}
& --
& --
& \third{0.020}{\tiny\,(2)}
& \textcolor{gray}{\hl{0.875}}
& \textcolor{gray}{\hl{0.796}}
& \textcolor{gray}{\hl{0.576}{\tiny\,(11)}}
& \textcolor{gray}{\hl{0.866}}
& \textcolor{gray}{--}
& \textcolor{gray}{\hl{0.411}{\tiny\,(3)}}
& --
& --
& \second{0.075}{\tiny\,(10)}
& \second{0.246}
& \second{0.216} \\

HistoPlexer~\cite{andani2025histoplexer} & -- & 0.011 & 0.003{\tiny\,(3)} & -- & -- & 0.005{\tiny\,(1)} & -- & -- & 0.004{\tiny\,(2)} & -- & -- & 0.007{\tiny\,(1)} & -- & -- & 0.002{\tiny\,(4)} & 0.003 & 0.006 \\
\midrule
\method & \hl{0.663} & \hl{0.542} & \hl{0.278}{\tiny\,(29)} & \hl{0.180} & \hl{0.154} & \hl{0.127}{\tiny\,(25)} & \second{0.805} & \second{0.723} & \second{0.449}{\tiny\,(11)} & \second{0.761} & \hl{0.243} & \second{0.356}{\tiny\,(4)} & \hl{0.834} & \hl{0.863} & \hl{0.584}{\tiny\,(57)} & \hl{0.437} & \hl{0.241} \\
\bottomrule
\end{tabular}
\end{table*}

%% file: tables/tab_ablation.tex
\begin{table*}[t]
\centering
\caption{\textbf{Joint training and fusion.} Comparison of the cell branch, dense branch,
and their fusion, trained jointly on RNA and protein
or separately on each modality.
Differences are computed as the first minus the second
method. $(n/m)$ indicates the number of evaluation units
where the first outperforms the second.}
\label{tab:ablation}
\footnotesize
\setlength{\tabcolsep}{7.1pt}
\begin{tabular}{@{}lcccccc@{}}
\toprule
 & \multicolumn{2}{c}{10x Genomics} & \multicolumn{2}{c}{CRC} & \multicolumn{2}{c}{ORION} \\
\cmidrule(lr){2-3}\cmidrule(lr){4-5}\cmidrule(l){6-7}
Model & Seen & Unseen & Seen & Unseen & Seen & Unseen \\
\midrule
\multicolumn{7}{@{}l}{\textit{Single-modality counterparts}} \\
\quad Cell branch & 0.155{\scriptsize$\pm$.000} & 0.094{\scriptsize$\pm$.005} & 0.211{\scriptsize$\pm$.001} & 0.073{\scriptsize$\pm$.010} & 0.398{\scriptsize$\pm$.007} & 0.203{\scriptsize$\pm$.018} \\
\quad Dense branch & \third{0.160}{\scriptsize$\pm$.001} & \third{0.107}{\scriptsize$\pm$.001} & 0.251{\scriptsize$\pm$.007} & \third{0.155}{\scriptsize$\pm$.014} & 0.424{\scriptsize$\pm$.007} & \hl{0.392}{\scriptsize$\pm$.008} \\
\quad Dual branch & \hl{0.172}{\scriptsize$\pm$.000} & \hl{0.118}{\scriptsize$\pm$.002} & \third{0.257}{\scriptsize$\pm$.003} & 0.144{\scriptsize$\pm$.008} & \third{0.447}{\scriptsize$\pm$.006} & \third{0.355}{\scriptsize$\pm$.006} \\
\multicolumn{7}{@{}l}{\textit{Joint (two-assay) training}} \\
\quad Cell branch & 0.151{\scriptsize$\pm$.000} & 0.089{\scriptsize$\pm$.004} & 0.238{\scriptsize$\pm$.001} & 0.066{\scriptsize$\pm$.008} & 0.422{\scriptsize$\pm$.003} & 0.270{\scriptsize$\pm$.032} \\
\quad Dense branch & 0.158{\scriptsize$\pm$.001} & \third{0.107}{\scriptsize$\pm$.003} & \second{0.272}{\scriptsize$\pm$.004} & \hl{0.225}{\scriptsize$\pm$.007} & \second{0.461}{\scriptsize$\pm$.014} & \second{0.386}{\scriptsize$\pm$.007} \\
\quad Dual branch & \second{0.170}{\scriptsize$\pm$.000} & \second{0.116}{\scriptsize$\pm$.003} & \hl{0.282}{\scriptsize$\pm$.002} & \second{0.185}{\scriptsize$\pm$.009} & \hl{0.476}{\scriptsize$\pm$.008} & \second{0.386}{\scriptsize$\pm$.010} \\
\midrule
\multicolumn{7}{@{}l}{\textit{Paired differences}} \\
\quad Joint cell $-$ Single cell & $-$0.0032 (1/16) & $-$0.0055 (2/16) & +0.0271 (34/35) & $-$0.0069 (12/35) & +0.0246 (40/40) & +0.0669 (37/40) \\
\quad Joint dense $-$ Single dense & $-$0.0021 (4/16) & +0.0004 (9/16) & +0.0209 (35/35) & +0.0698 (35/35) & +0.0373 (40/40) & $-$0.0062 (18/40) \\
\quad Joint dual $-$ Single dual & $-$0.0022 (1/16) & $-$0.0024 (5/16) & +0.0244 (35/35) & +0.0413 (35/35) & +0.0288 (40/40) & +0.0306 (32/40) \\
\quad Joint dual $-$ Joint cell & +0.0187 (16/16) & +0.0269 (16/16) & +0.0439 (35/35) & +0.1191 (35/35) & +0.0534 (40/40) & +0.1159 (40/40) \\
\quad Joint dual $-$ Joint dense & +0.0120 (12/16) & +0.0086 (9/16) & +0.0099 (31/35) & $-$0.0397 (0/35) & +0.0144 (38/40) & +0.0003 (22/40) \\
\bottomrule
\end{tabular}
\end{table*}

%% file: tables/tab_knowledge.tex
\begin{table}[!t]
\centering
\caption{\textbf{Gene embeddings on markers without \HE labels.} Each variant retrains both branches.
Remove $\mathbf{T}^0$ replaces initial embeddings
with scale-matched random vectors.
Remove $\mathbf{S}$ disables relation-based composition.
No knowledge removes both.
RNA evaluates 98 withheld genes with relation weights.
Protein evaluates 23 CRC and 3 ORION unseen markers.
Results are mean $\pm$ s.d.\ over 3 seeds. }
\label{tab:knowledge}
\footnotesize
\setlength{\tabcolsep}{6pt}
\begin{tabular}{@{}lccc@{}}
\toprule
 & Withheld & \multicolumn{2}{c}{Unseen protein} \\
\cmidrule(lr){3-4}
Gene embeddings & 10x Genomics (98) & CRC (23) & ORION (3) \\
\midrule
No knowledge & 0.040{\scriptsize$\pm$.002} & \third{0.152}{\scriptsize$\pm$.002} & 0.176{\scriptsize$\pm$.049} \\
Remove $\mathbf{T}^0$ & \third{0.083}{\scriptsize$\pm$.006} & 0.141{\scriptsize$\pm$.011} & \third{0.237}{\scriptsize$\pm$.118} \\
Remove $\mathbf{S}$ & \second{0.105}{\scriptsize$\pm$.007} & \second{0.175}{\scriptsize$\pm$.010} & \hl{0.387}{\scriptsize$\pm$.017} \\
\method & \hl{0.129}{\scriptsize$\pm$.004} & \hl{0.185}{\scriptsize$\pm$.011} & \second{0.386}{\scriptsize$\pm$.012} \\
\bottomrule
\end{tabular}
\end{table}

%% file: sec/6_conclusion.tex
\section{Conclusion}
\label{sec:conclusion}

We presented \method, which learns to predict the RNA and protein of every cell from \HE alone using unpaired sections, each measuring one assay with its own panel.
One learnable embedding per gene, read by both the RNA and the protein heads and composed from the embeddings of related genes, connects the two assays and reaches markers that no section labels.

Three findings follow. First, RNA supervision improves protein prediction on both external cohorts, although no section measures both assays. Second, gene embeddings composed from related genes predict RNA genes that no \HE section labels, and the gain depends on the right partners. Third, an equal-weight fusion of the cell-level branch and the dense branch beats the cell-level branch on every test set and the dense branch on the seen markers of every test set. Across RNA and protein, \method outperforms the
strongest RNA baseline on average and, averaged over the two external cohorts, the strongest protein baseline on seen and unseen markers.

%% file: main.bbl
\begin{thebibliography}{40}
\providecommand{\natexlab}[1]{#1}
\providecommand{\url}[1]{\texttt{#1}}
\expandafter\ifx\csname urlstyle\endcsname\relax
  \providecommand{\doi}[1]{doi: #1}\else
  \providecommand{\doi}{doi: \begingroup \urlstyle{rm}\Url}\fi

\bibitem[Andani et~al.(2025)Andani, Chen, Ficek-Pascual, Heinke, Casanova, Hild, Sobottka, Bodenmiller, {Tumor Profiler Consortium}, Koelzer, and R{\"a}tsch]{andani2025histoplexer}
Sonali Andani, Boqi Chen, Joanna Ficek-Pascual, Simon Heinke, Ruben Casanova, Bernard~Friedrich Hild, Bettina Sobottka, Bernd Bodenmiller, {Tumor Profiler Consortium}, Viktor~H. Koelzer, and Gunnar R{\"a}tsch.
\newblock Histopathology-based protein multiplex generation using deep learning.
\newblock \emph{Nature Machine Intelligence}, 7\penalty0 (8):\penalty0 1292--1307, 2025.

\bibitem[Balezo et~al.(2026)Balezo, Trullo, Pla~Planas, Decenci{\`e}re, and Walter]{balezo2025miphei}
Guillaume Balezo, Roger Trullo, Albert Pla~Planas, Etienne Decenci{\`e}re, and Thomas Walter.
\newblock {MIPHEI-ViT}: Multiplex immunofluorescence prediction from {H\&E} images using {ViT} foundation models.
\newblock \emph{Computers in Biology and Medicine}, 206:\penalty0 111564, 2026.

\bibitem[Belkin and Niyogi(2002)]{belkin2002laplacian}
Mikhail Belkin and Partha Niyogi.
\newblock Laplacian eigenmaps and spectral techniques for embedding and clustering.
\newblock In \emph{Advances in Neural Information Processing Systems 14}, pages 585--592. MIT Press, 2002.

\bibitem[Chadoutaud et~al.(2026)Chadoutaud, Lerousseau, Herrero-Saboya, Ostermaier, Fontugne, Barillot, and Walter]{chadoutaud2026scellst}
Lo{\"i}c Chadoutaud, Marvin Lerousseau, Daniel Herrero-Saboya, Julian Ostermaier, Jacqueline Fontugne, Emmanuel Barillot, and Thomas Walter.
\newblock {sCellST} predicts single-cell gene expression from {H\&E} images.
\newblock \emph{Nature Communications}, 17:\penalty0 1194, 2026.

\bibitem[Dezem et~al.(2024)Dezem, Arjumand, DuBose, Morosini, and Plummer]{dezem2024spatially}
Felipe~Segato Dezem, Wani Arjumand, Hannah DuBose, Natalia~Silva Morosini, and Jasmine Plummer.
\newblock Spatially resolved single-cell omics: Methods, challenges, and future perspectives.
\newblock \emph{Annual Review of Biomedical Data Science}, 7\penalty0 (1):\penalty0 131--153, 2024.

\bibitem[Fang et~al.(2026)Fang, Li, Wang, Zeng, and Min]{fang2026hinge}
Donghai Fang, Yongheng Li, Zhen Wang, Yuansong Zeng, and Wenwen Min.
\newblock Adapting a pre-trained single-cell foundation model to spatial gene expression generation from histology images.
\newblock In \emph{Proceedings of the IEEE/CVF Conference on Computer Vision and Pattern Recognition (CVPR)}, pages 5720--5729, 2026.

\bibitem[Gao et~al.(2026)Gao, Gu, Xiang, Luo, and Heng]{gao2026cello}
Zijun Gao, Chunbin Gu, Jinxi Xiang, Xiangde Luo, and Pheng-Ann Heng.
\newblock Towards scalable context-aware single-cell spatial transcriptomics prediction from histology images.
\newblock \emph{arXiv preprint arXiv:2609.36429}, 2026.
\newblock Accepted to NeurIPS 2026.

\bibitem[Gatenbee et~al.(2023)Gatenbee, Baker, Prabhakaran, Swinyard, Slebos, Mandal, Mulholland, Andor, Marusyk, Leedham, Conejo-Garcia, Chung, Robertson-Tessi, Graham, and Anderson]{gatenbee2023valis}
Chandler~D. Gatenbee, Ann-Marie Baker, Sandhya Prabhakaran, Ottilie Swinyard, Robbert J.~C. Slebos, Gunjan Mandal, Eoghan Mulholland, Noemi Andor, Andriy Marusyk, Simon Leedham, Jose~R. Conejo-Garcia, Christine~H. Chung, Mark Robertson-Tessi, Trevor~A. Graham, and Alexander R.~A. Anderson.
\newblock Virtual alignment of pathology image series for multi-gigapixel whole slide images.
\newblock \emph{Nature Communications}, 14:\penalty0 4502, 2023.

\bibitem[Ge and Cai(2026)]{ge2026cube}
Zhiyan Ge and Haoyang Cai.
\newblock {CUBE}: Multimodal representation learning reveals biological structure across histomorphology, spatial protein phenotypes, and transcriptome-associated signals.
\newblock \emph{bioRxiv}, 2026.

\bibitem[Goltsev et~al.(2018)Goltsev, Samusik, Kennedy-Darling, Bhate, Hale, Vazquez, Black, and Nolan]{goltsev2018codex}
Yury Goltsev, Nikolay Samusik, Julia Kennedy-Darling, Salil Bhate, Matthew Hale, Gustavo Vazquez, Sarah Black, and Garry~P. Nolan.
\newblock Deep profiling of mouse splenic architecture with {CODEX} multiplexed imaging.
\newblock \emph{Cell}, 174\penalty0 (4):\penalty0 968--981.e15, 2018.

\bibitem[Gu et~al.(2026)Gu, Wu, Yan, Wang, Li, Lin, Cui, Lai, Luo, Zhou, Yuan, and Yao]{gu2026h2o}
Yunjie Gu, Zihan Wu, Rui Yan, Zhikang Wang, Yuan Li, Senlin Lin, Yan Cui, Haoran Lai, Xin Luo, Shaohua~Kevin Zhou, Zhiyuan Yuan, and Jianhua Yao.
\newblock {H2O}: A foundation model bridging histopathology to spatial multi-omics profiling.
\newblock \emph{bioRxiv}, 2026.

\bibitem[He et~al.(2020)He, Bergenstr{\aa}hle, Stenbeck, Abid, Andersson, Borg, Maaskola, Lundeberg, and Zou]{he2020stnet}
Bryan He, Ludvig Bergenstr{\aa}hle, Linnea Stenbeck, Abubakar Abid, Alma Andersson, {\AA}ke Borg, Jonas Maaskola, Joakim Lundeberg, and James Zou.
\newblock Integrating spatial gene expression and breast tumour morphology via deep learning.
\newblock \emph{Nature Biomedical Engineering}, 4\penalty0 (8):\penalty0 827--834, 2020.

\bibitem[Hickey et~al.(2022)Hickey, Neumann, Radtke, Camarillo, Beuschel, Albanese, McDonough, Hatler, Wiblin, Fisher, Croteau, Small, Sood, Caprioli, Angelo, Nolan, Chung, Hewitt, Germain, Spraggins, Lundberg, Snyder, Kelleher, and Saka]{hickey2022spatial}
John~W. Hickey, Elizabeth~K. Neumann, Andrea~J. Radtke, Jeannie~M. Camarillo, Rebecca~T. Beuschel, Alexandre Albanese, Elizabeth McDonough, Julia Hatler, Anne~E. Wiblin, Jeremy Fisher, Josh Croteau, Eliza~C. Small, Anup Sood, Richard~M. Caprioli, R.~Michael Angelo, Garry~P. Nolan, Kwanghun Chung, Stephen~M. Hewitt, Ronald~N. Germain, Jeffrey~M. Spraggins, Emma Lundberg, Michael~P. Snyder, Neil~L. Kelleher, and Sinem~K. Saka.
\newblock Spatial mapping of protein composition and tissue organization: a primer for multiplexed antibody-based imaging.
\newblock \emph{Nature Methods}, 19\penalty0 (3):\penalty0 284--295, 2022.

\bibitem[Janesick et~al.(2023)Janesick, Shelansky, Gottscho, Wagner, Williams, Rouault, Beliakoff, Morrison, Oliveira, Sicherman, Kohlway, Abousoud, Drennon, Mohabbat, {10x Development Teams}, and Taylor]{janesick2023xenium}
Amanda Janesick, Robert Shelansky, Andrew~D. Gottscho, Florian Wagner, Stephen~R. Williams, Morgane Rouault, Ghezal Beliakoff, Carolyn~A. Morrison, Michelli~F. Oliveira, Jordan~T. Sicherman, Andrew Kohlway, Jawad Abousoud, Tingsheng~Yu Drennon, Seayar~H. Mohabbat, {10x Development Teams}, and Sarah E.~B. Taylor.
\newblock High resolution mapping of the tumor microenvironment using integrated single-cell, spatial and in situ analysis.
\newblock \emph{Nature Communications}, 14\penalty0 (1):\penalty0 8353, 2023.

\bibitem[Ji et~al.(2026)Ji, Hu, Wang, Liu, Xu, Zhang, Zhong, Qiao, Zhang, Peng, and Yu]{ji2026captain}
Boya Ji, Tingting Hu, Jiawen Wang, Mengmeng Liu, Liwen Xu, Qinhao Zhang, Siyun Zhong, Libo Qiao, Yan Zhang, Shaoliang Peng, and Fulong Yu.
\newblock {CAPTAIN}: a multimodal foundation model pretrained on co-assayed single-cell {RNA} and protein.
\newblock \emph{Nature Communications}, 17:\penalty0 6161, 2026.

\bibitem[Lin et~al.(2023{\natexlab{a}})Lin, Chen, Campton, Cooper, Coy, Yapp, Tefft, McCarty, Ligon, Rodig, Reese, George, Santagata, and Sorger]{lin2023orion}
Jia-Ren Lin, Yu-An Chen, Daniel Campton, Jeremy Cooper, Shannon Coy, Clarence Yapp, Juliann~B. Tefft, Erin McCarty, Keith~L. Ligon, Scott~J. Rodig, Steven Reese, Tad George, Sandro Santagata, and Peter~K. Sorger.
\newblock High-plex immunofluorescence imaging and traditional histology of the same tissue section for discovering image-based biomarkers.
\newblock \emph{Nature Cancer}, 4:\penalty0 1036--1052, 2023{\natexlab{a}}.

\bibitem[Lin et~al.(2023{\natexlab{b}})Lin, Akin, Rao, Hie, Zhu, Lu, Smetanin, Verkuil, Kabeli, Shmueli, dos Santos~Costa, Fazel-Zarandi, Sercu, Candido, and Rives]{lin2023esm2}
Zeming Lin, Halil Akin, Roshan Rao, Brian Hie, Zhongkai Zhu, Wenting Lu, Nikita Smetanin, Robert Verkuil, Ori Kabeli, Yaniv Shmueli, Allan dos Santos~Costa, Maryam Fazel-Zarandi, Tom Sercu, Salvatore Candido, and Alexander Rives.
\newblock Evolutionary-scale prediction of atomic-level protein structure with a language model.
\newblock \emph{Science}, 379\penalty0 (6637):\penalty0 1123--1130, 2023{\natexlab{b}}.

\bibitem[Liu et~al.(2022)Liu, Mao, Wu, Feichtenhofer, Darrell, and Xie]{liu2022convnext}
Zhuang Liu, Hanzi Mao, Chao-Yuan Wu, Christoph Feichtenhofer, Trevor Darrell, and Saining Xie.
\newblock A {ConvNet} for the 2020s.
\newblock In \emph{Proceedings of the IEEE/CVF Conference on Computer Vision and Pattern Recognition (CVPR)}, pages 11976--11986, 2022.

\bibitem[Loshchilov and Hutter(2019)]{loshchilov2019adamw}
Ilya Loshchilov and Frank Hutter.
\newblock Decoupled weight decay regularization.
\newblock In \emph{International Conference on Learning Representations (ICLR)}, 2019.

\bibitem[Milacic et~al.(2024)Milacic, Beavers, Conley, Gong, Gillespie, Griss, Haw, Jassal, Matthews, May, Petryszak, Ragueneau, Rothfels, Sevilla, Shamovsky, Stephan, Tiwari, Varusai, Weiser, Wright, Wu, Stein, Hermjakob, and D'Eustachio]{milacic2024reactome}
Marija Milacic, Deidre Beavers, Patrick Conley, Chuqiao Gong, Marc Gillespie, Johannes Griss, Robin Haw, Bijay Jassal, Lisa Matthews, Bruce May, Robert Petryszak, Eliot Ragueneau, Karen Rothfels, Cristoffer Sevilla, Veronica Shamovsky, Ralf Stephan, Krishna Tiwari, Thawfeek Varusai, Joel Weiser, Adam Wright, Guanming Wu, Lincoln Stein, Henning Hermjakob, and Peter D'Eustachio.
\newblock The {Reactome} pathway knowledgebase 2024.
\newblock \emph{Nucleic Acids Research}, 52\penalty0 (D1):\penalty0 D672--D678, 2024.

\bibitem[Nonchev et~al.(2025)Nonchev, Manaiev, Koelzer, and R{\"a}tsch]{nonchev2025deepspot2cell}
Kalin Nonchev, Glib Manaiev, Viktor~H. Koelzer, and Gunnar R{\"a}tsch.
\newblock {DeepSpot2Cell}: Predicting virtual single-cell spatial transcriptomics from {H\&E} images using spot-level supervision.
\newblock \emph{bioRxiv}, 2025.

\bibitem[Nonchev et~al.(2026)Nonchev, Dawo, Silina, Koelzer, and R{\"a}tsch]{nonchev2026deepspotm}
Kalin Nonchev, Sebastian Dawo, Karina Silina, Viktor~H. Koelzer, and Gunnar R{\"a}tsch.
\newblock {DeepSpot-M}: a multimodal foundation model for transcriptome-wide virtual spatial transcriptomics from histology.
\newblock \emph{medRxiv}, 2026.

\bibitem[Rao et~al.(2021)Rao, Barkley, Fran{\c{c}}a, and Yanai]{rao2021exploring}
Anjali Rao, Dalia Barkley, Gustavo~S. Fran{\c{c}}a, and Itai Yanai.
\newblock Exploring tissue architecture using spatial transcriptomics.
\newblock \emph{Nature}, 596\penalty0 (7871):\penalty0 211--220, 2021.

\bibitem[Rosen et~al.(2026)Rosen, Roohani, Agrawal, Samotor{\v{c}}an, {Tabula Sapiens Consortium}, Quake, and Leskovec]{rosen2026uce}
Yanay Rosen, Yusuf Roohani, Ayush Agrawal, Leon Samotor{\v{c}}an, {Tabula Sapiens Consortium}, Stephen~R. Quake, and Jure Leskovec.
\newblock Universal cell embedding provides a foundation model for cell biology.
\newblock \emph{Nature}, 656\penalty0 (8126):\penalty0 183--191, 2026.

\bibitem[Scalbert et~al.(2026)Scalbert, Saillard, Peeters, Gonzalez, Valter, Llinares-L{\'o}pez, Mariet, and Jenatton]{scalbert2026hoptimus1}
Marin Scalbert, Charlie Saillard, Thomas Peeters, Liam Gonzalez, Dasha Valter, Felipe Llinares-L{\'o}pez, Zelda~E. Mariet, and Rodolphe Jenatton.
\newblock {H-optimus-1}: A foundation model for computational histopathology.
\newblock In \emph{Proceedings of the American Association for Cancer Research Annual Meeting 2026; Part 2 (Late-Breaking, Clinical Trial, and Invited Abstracts)}, page LB174. AACR, 2026.

\bibitem[Sch{\"u}rch et~al.(2020{\natexlab{a}})Sch{\"u}rch, Bhate, Barlow, Phillips, Noti, Zlobec, Chu, Black, Demeter, McIlwain, Samusik, Goltsev, and Nolan]{schurch2020crcdata}
Christian~M. Sch{\"u}rch, Salil Bhate, Graham Barlow, Darci Phillips, Luca Noti, Inti Zlobec, Pauline Chu, Sarah Black, Janos Demeter, David McIlwain, Nikolay Samusik, Yury Goltsev, and Garry Nolan.
\newblock High-dimensional imaging of colorectal carcinoma and other tumors with 50+ markers.
\newblock The Cancer Imaging Archive, \url{https://doi.org/10.7937/TCIA.2020.FQN0-0326}, 2020{\natexlab{a}}.

\bibitem[Sch{\"u}rch et~al.(2020{\natexlab{b}})Sch{\"u}rch, Bhate, Barlow, Phillips, Noti, Zlobec, Chu, Black, Demeter, McIlwain, Kinoshita, Samusik, Goltsev, and Nolan]{schurch2020crc}
Christian~M. Sch{\"u}rch, Salil~S. Bhate, Graham~L. Barlow, Darci~J. Phillips, Luca Noti, Inti Zlobec, Pauline Chu, Sarah Black, Janos Demeter, David~R. McIlwain, Shigemi Kinoshita, Nikolay Samusik, Yury Goltsev, and Garry~P. Nolan.
\newblock Coordinated cellular neighborhoods orchestrate antitumoral immunity at the colorectal cancer invasive front.
\newblock \emph{Cell}, 182\penalty0 (5):\penalty0 1341--1359.e19, 2020{\natexlab{b}}.

\bibitem[Stoeckius et~al.(2017)Stoeckius, Hafemeister, Stephenson, Houck-Loomis, Chattopadhyay, Swerdlow, Satija, and Smibert]{stoeckius2017citeseq}
Marlon Stoeckius, Christoph Hafemeister, William Stephenson, Brian Houck-Loomis, Pratip~K. Chattopadhyay, Harold Swerdlow, Rahul Satija, and Peter Smibert.
\newblock Simultaneous epitope and transcriptome measurement in single cells.
\newblock \emph{Nature Methods}, 14\penalty0 (9):\penalty0 865--868, 2017.

\bibitem[Szklarczyk et~al.(2023)Szklarczyk, Kirsch, Koutrouli, Nastou, Mehryary, Hachilif, Gable, Fang, Doncheva, Pyysalo, Bork, Jensen, and von Mering]{szklarczyk2023string}
Damian Szklarczyk, Rebecca Kirsch, Mikaela Koutrouli, Katerina Nastou, Farrokh Mehryary, Radja Hachilif, Annika~L. Gable, Tao Fang, Nadezhda~T. Doncheva, Sampo Pyysalo, Peer Bork, Lars~J. Jensen, and Christian von Mering.
\newblock The {STRING} database in 2023: protein--protein association networks and functional enrichment analyses for any sequenced genome of interest.
\newblock \emph{Nucleic Acids Research}, 51\penalty0 (D1):\penalty0 D638--D646, 2023.

\bibitem[Tang et~al.(2023)Tang, Wen, Liu, Ding, Jin, Xie, Liu, and Tang]{tang2023scmoformer}
Wenzhuo Tang, Hongzhi Wen, Renming Liu, Jiayuan Ding, Wei Jin, Yuying Xie, Hui Liu, and Jiliang Tang.
\newblock Single-cell multimodal prediction via transformers.
\newblock In \emph{Proceedings of the 32nd ACM International Conference on Information and Knowledge Management (CIKM)}, pages 2422--2431, 2023.

\bibitem[Valanarasu et~al.(2026)Valanarasu, Xu, Usuyama, Kim, Wong, Argaw, Ben~Shimol, Crabtree, Matlock, Bartlett, Bagga, Gu, Zhang, Naumann, Fox, Wright, Robicsek, Piening, Bifulco, Wang, and Poon]{valanarasu2026gigatime}
Jeya Maria~Jose Valanarasu, Hanwen Xu, Naoto Usuyama, Chanwoo Kim, Cliff Wong, Peniel Argaw, Racheli Ben~Shimol, Angela Crabtree, Kevin Matlock, Alexandra~Q. Bartlett, Jaspreet Bagga, Yu Gu, Sheng Zhang, Tristan Naumann, Bernard~A. Fox, Bill Wright, Ari Robicsek, Brian Piening, Carlo Bifulco, Sheng Wang, and Hoifung Poon.
\newblock Multimodal {AI} generates virtual population for tumor microenvironment modeling.
\newblock \emph{Cell}, 189\penalty0 (2):\penalty0 386--400.e19, 2026.

\bibitem[Walter et~al.(2023)Walter, Stegle, and Velten]{walter2023fishfactor}
Florin~C. Walter, Oliver Stegle, and Britta Velten.
\newblock {FISHFactor}: a probabilistic factor model for spatial transcriptomics data with subcellular resolution.
\newblock \emph{Bioinformatics}, 39\penalty0 (5):\penalty0 btad183, 2023.

\bibitem[Wenckstern et~al.(2026)Wenckstern, Jain, von Querfurth, Cheng, Vasilev, Pariset, Cheng, Liakopoulos, Michielin, Wicki, Gut, and Bunne]{wenckstern2026virtues}
Johann Wenckstern, Eeshaan Jain, Benedikt von Querfurth, Yexiang Cheng, Kiril Vasilev, Matteo Pariset, Phil~F. Cheng, Petros Liakopoulos, Olivier Michielin, Andreas Wicki, Gabriele Gut, and Charlotte Bunne.
\newblock The {Virtual Tissues} foundation model resolves spatial proteomics across scales.
\newblock \emph{Nature}, 657\penalty0 (8132):\penalty0 744--754, 2026.

\bibitem[Wu et~al.(2026)Wu, Zuo, and Xie]{wu2026gsatrag}
Chaochen Wu, Meiyun Zuo, and Lei Xie.
\newblock Open-set spatial gene expression prediction from histological images via retrieval-augmented generation.
\newblock In \emph{Proceedings of the IEEE/CVF Conference on Computer Vision and Pattern Recognition (CVPR) Findings}, pages 5495--5504, 2026.

\bibitem[Wu et~al.(2025)Wu, Bieniosek, Wu, Thakkar, Charville, Makky, Sch{\"u}rch, Huyghe, Peters, Li, Li, Giba, Behera, Raman, Trevino, Mayer, and Zou]{wu2025rosie}
Eric Wu, Matthew Bieniosek, Zhenqin Wu, Nitya Thakkar, Gregory~W. Charville, Ahmad Makky, Christian~M. Sch{\"u}rch, Jeroen~R. Huyghe, Ulrike Peters, Christopher~I. Li, Li Li, Hannah Giba, Vivek Behera, Arjun Raman, Alexandro~E. Trevino, Aaron~T. Mayer, and James Zou.
\newblock {ROSIE}: {AI} generation of multiplex immunofluorescence staining from histopathology images.
\newblock \emph{Nature Communications}, 16:\penalty0 7633, 2025.

\bibitem[Yang et~al.(2023)Yang, Pan, Wang, Yu, Shen, Chen, Xiao, Jiang, and Guo]{yang2023adatask}
Enneng Yang, Junwei Pan, Ximei Wang, Haibin Yu, Li Shen, Xihua Chen, Lei Xiao, Jie Jiang, and Guibing Guo.
\newblock {AdaTask}: A task-aware adaptive learning rate approach to multi-task learning.
\newblock In \emph{Proceedings of the AAAI Conference on Artificial Intelligence}, pages 10745--10753, 2023.

\bibitem[Yang et~al.(2024)Yang, Hossain, Li, Rahman, and Stone]{yang2024sgn}
Yan Yang, Md~Zakir Hossain, Xuesong Li, Shafin Rahman, and Eric Stone.
\newblock Spatial transcriptomics analysis of zero-shot gene expression prediction.
\newblock In \emph{Medical Image Computing and Computer Assisted Intervention -- {MICCAI} 2024}, pages 492--502. Springer, 2024.

\bibitem[Yao et~al.(2026)Yao, Schroeder, Jiang, Im, Park, Dumoulin, Hwang, Susztak, and Li]{yao2026pixel2gene}
Sicong Yao, Amelia Schroeder, Shunzhou Jiang, Soyoung Im, Jeong~Hwan Park, Bernhard Dumoulin, Tae~Hyun Hwang, Katalin Susztak, and Mingyao Li.
\newblock {Pixel2Gene} enables histology-guided reconstruction and prediction of spatial gene expression.
\newblock \emph{bioRxiv}, 2026.

\bibitem[Zhang et~al.(2024)Zhang, Schroeder, Yan, Yang, Hu, Lee, Cho, Susztak, Xu, Feldman, Lee, Furth, Wang, and Li]{zhang2024istar}
Daiwei Zhang, Amelia Schroeder, Hanying Yan, Haochen Yang, Jian Hu, Michelle Y.~Y. Lee, Kyung~S. Cho, Katalin Susztak, George~X. Xu, Michael~D. Feldman, Edward~B. Lee, Emma~E. Furth, Linghua Wang, and Mingyao Li.
\newblock Inferring super-resolution tissue architecture by integrating spatial transcriptomics with histology.
\newblock \emph{Nature Biotechnology}, 42\penalty0 (9):\penalty0 1372--1377, 2024.

\bibitem[Zhang et~al.(2026)Zhang, Gong, Pang, Liu, Lu, Cui, Ge, Zeng, Yi, Li, Liu, Yu, Wang, Crispin-Ortuzar, Yu, Li, and Gao]{zhang2026care}
Di Zhang, Zhangpeng Gong, Xiaobo Pang, Jiashuai Liu, Junbo Lu, Hao Cui, Jiusong Ge, Zhi Zeng, Kai Yi, Yinghua Li, Si Liu, Tingsong Yu, Haoran Wang, Mireia Crispin-Ortuzar, Weimiao Yu, Chen Li, and Zeyu Gao.
\newblock {CARE}: A molecular-guided foundation model with adaptive region modeling for whole slide image analysis.
\newblock In \emph{Proceedings of the IEEE/CVF Conference on Computer Vision and Pattern Recognition (CVPR)}, pages 21078--21088, 2026.

\end{thebibliography}
